\documentclass[conference]{IEEEtran}
\IEEEoverridecommandlockouts
\usepackage[hidelinks]{hyperref}
\usepackage{cite}
\usepackage{amsmath,amssymb,amsfonts}
\usepackage{algorithmic}
\usepackage{graphicx}
\usepackage{textcomp}
\usepackage{xcolor}
\def\BibTeX{{\rm B\kern-.05em{\sc i\kern-.025em b}\kern-.08em
    T\kern-.1667em\lower.7ex\hbox{E}\kern-.125emX}}

\usepackage{verbatim}
\usepackage{booktabs} % table top, middle, bottom rules
\newcommand{\martinTableFontsize}{\fontsize{7.4}{7.4}\selectfont}
\newcommand{\MSCS}{MPMCS}
\newcommand{\toolname}{MPMCS4FTA}
\newcommand{\martinListingFontsize}{\fontsize{9}{9}\selectfont}

\newcommand{\ackContent}{
    This work has been supported by the European Union's Horizon 2020 research and innovation programme under grant %agreement 
    No 739551 (KIOS CoE). 
}

\newcommand{\confname}{Accepted for publication at the 50th IEEE/IFIP International Conference on Dependable Systems and Networks (\textbf{DSN 2020}), Fast Abstracts Track. \copyright2020 IEEE.}
\makeatletter
\def\ps@IEEEtitlepagestyle{%
    \def\@oddfoot{\mycopyrightnotice}%
    \def\@evenfoot{}%
}
\def\mycopyrightnotice{%
    {\footnotesize \confname}
    \gdef\mycopyrightnotice{}
}

\begin{document}

\title{Fault Tree Analysis: Identifying Maximum Probability Minimal Cut Sets with MaxSAT
    \vspace{-0.28cm}
\thanks{\ackContent}
}

\author{\IEEEauthorblockN{Mart\'in Barr\`ere and Chris Hankin
    }
    \IEEEauthorblockA{
    Institute for Security Science and Technology, Imperial College London, UK\\
    \{m.barrere, c.hankin\}@imperial.ac.uk}
}

\maketitle
\pagestyle{plain}

\begin{abstract}
In this paper, we present a novel MaxSAT-based technique to compute Maximum Probability Minimal Cut Sets (\MSCS s) in fault trees. 
We model the \MSCS~problem as a Weighted Partial MaxSAT problem and solve it using a parallel SAT-solving architecture. 
The results obtained with our open source tool indicate that the approach is effective and efficient. 
\end{abstract}


\vspace{-0.5cm}
\section{Introduction}
\vspace{-0.05cm}
Fault Tree Analysis (FTA) constitutes a fundamental analytical tool aimed at modelling and evaluating how complex systems may fail \cite{FtaHandbook2002}. FTA is widely used in safety and reliability engineering as a risk assessment tool for a variety of industries such as aerospace, power plants, nuclear plants, and other high-hazard fields \cite{Ruijters2015}. Essentially, a fault tree (FT) involves a set of \textit{basic events} that are combined using logic operators (e.g. AND and OR gates) in order to model how these events may lead to an undesired system state represented at the root of the tree (top event). Basic events can be associated to hardware failures, human errors, and other cyber-physical conditions including cyber events such as software errors, communication failures, and cyber attacks. Let us consider a simple example. 

\vspace{-0.05cm}
\subsection{Fault tree example}
\begin{figure}[!t]
    \centering    
    \includegraphics[width=0.46\textwidth]{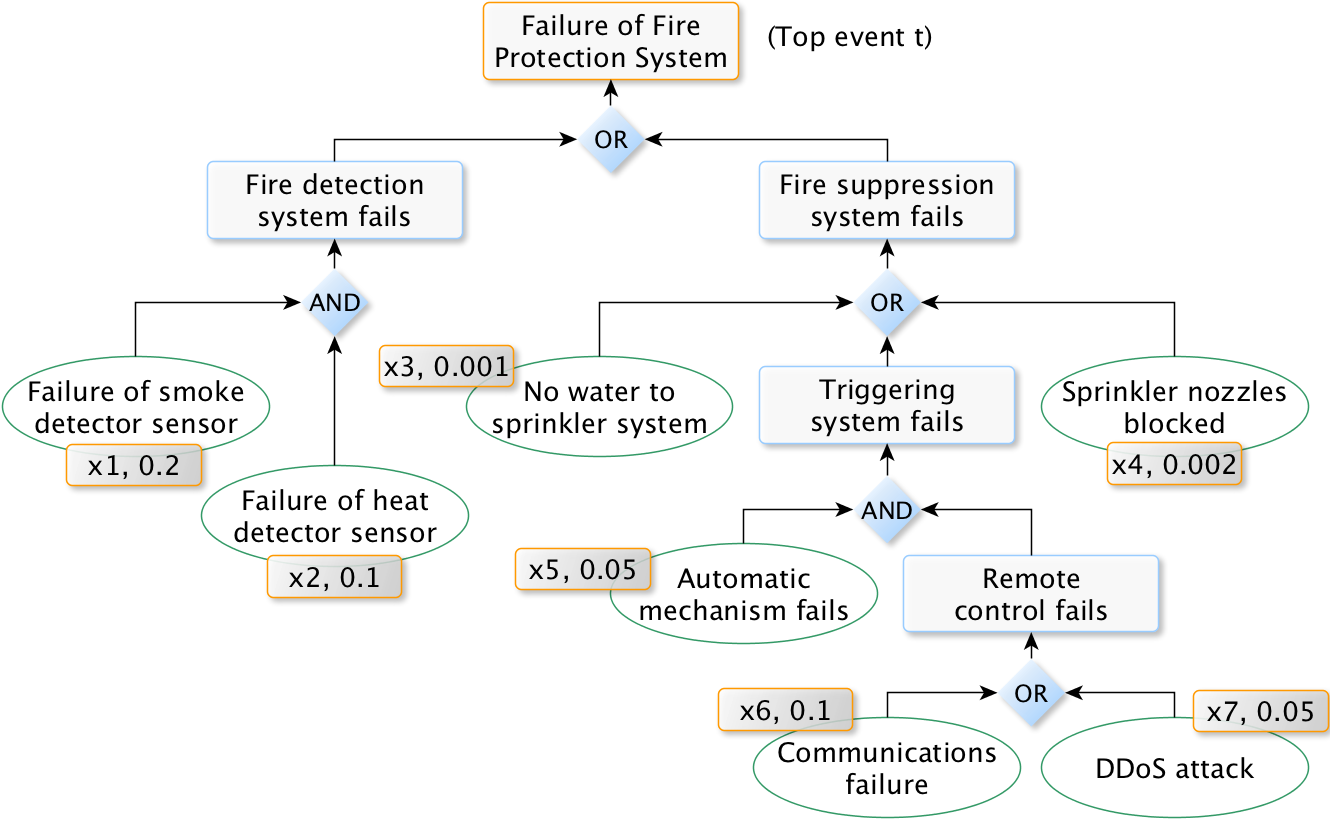}   
    \vspace{-0.2cm}
    \caption{Fault tree of a cyber-physical fire protection system (simplified)} 
    \label{fig:simple-example}
    \vspace{-0.6cm}
\end{figure}

\vspace{-0.05cm}
The fault tree shown in Fig. \ref{fig:simple-example} illustrates the different combinations of events that may lead to the failure of an hypothetical Fire Protection System (FPS) based on \cite{Kabir2017}. The FPS can fail if either the fire detection system or the fire suppression mechanism fails. In turn, the detection system can fail if both sensors fail simultaneously (events $x_1$ and $x_2$), while the suppression mechanism may fail if there is no water ($x3$), the sprinkler nozzles are blocked ($x_4$), or the triggering system does not work. The latter can fail if neither of its operation modes (automatic ($x_5$) or remotely operated) works properly. The remote control can fail if the communications channel fails ($x_6$) or the channel is not available due to a cyber attack, e.g. DDoS attack ($x_7$). Each basic event has an associated value that indicates its probability of occurrence~$p(x_i)$. 

\section{Problem description}
FTA comprises a broad family of methods and techniques used for qualitative and quantitative analysis. Qualitative techniques normally involve structural aspects of faults trees like single points of failure (SPOFs) and minimal cut sets (MCSs). MCSs are minimal combinations of events that together may lead to the failure of the top level event \cite{Ruijters2015}. 
Quantitative analysis usually involves numerical outcomes such as failure probabilities. The present work lies in the intersection of these two families. On the one hand, we are interested in finding MCSs. On the other hand, we focus on the MCS whose probability is the highest among all possible MCSs. 
We call this MCS the \textit{Maximum Probability Minimal Cut Set} (\MSCS). 
Note that this optimisation problem not only relates to the structural minimal cut set in the fault tree but also to the probabilities assigned to the events in it. 

A fault tree $F$ can be represented as a Boolean equation $f(t)$ that expresses the different ways in which the top event $t$ can be satisfied \cite{FtaHandbook2002}. 
In our example, $f(t)$ is as follows: 

\vspace{-0.15cm}
\martinListingFontsize\sffamily{ 
    \begin{equation}
    \begin{array}{c}
    f(t) = (x_1 \land x_2) \lor (x_3 \lor x_4 \lor (x_5 \land (x_6 \lor x_7))) \nonumber
    \end{array} 
      \vspace{-0.15cm}
    \end{equation}
}\normalsize \normalfont

The objective is to find the minimal set of logical variables that makes the equation $f(t)$ \textit{true} and whose joint probability is maximal among all minimal sets. 
In our example, the \MSCS~is $\{x_1,x_2\}$ with a joint probability of $0.02$. 

\section{Resolution method}
Our resolution method relies on Maximum Satisfiability (MaxSAT) techniques~\cite{Barrere-Arxiv2019}. A MaxSAT problem consists in finding a truth assignment that maximises the weight of the satisfied clauses. Equivalently, MaxSAT minimises the weight of the clauses it falsifies \cite{Davies2011}. A Weighted Partial MaxSAT problem involves \textit{soft clauses} with non-unit weights and it will try to minimise the penalty induced by falsified weighted variables. We use this last variant to solve our optimisation problem. The proposed resolution method involves six steps.

\textbf{Step 1 (Logical transformation).}
Since MaxSAT tries to maximise the number of satisfied clauses, and we are looking for minimal cut sets, we consider the complement of the equation $f(t)$ to represent the non-occurrence of the top event (system success): $X(t) = \neg f(t)$. 
$X(t)$ models the \textit{Success Tree} and can be obtained directly from the original fault tree by complementing all the events and substituting OR gates by AND gates, and vice versa~\cite{FtaHandbook2002}: 

\vspace{-0.4cm}
\martinListingFontsize\sffamily{ 
    \begin{equation}
    \begin{array}{c}
    X(t) = (\neg x_1 \lor \neg x_2) \land (\neg x_3 \land \neg x_4 \land (\neg x_5 \lor (\neg x_6 \land \neg x_7))) \nonumber
    \end{array} 
    \vspace{-0.2cm}
    \end{equation}
}\normalsize \normalfont

Since we are interested in minimising the number of satisfied clauses, which is opposed to what MaxSAT does (maximisation), we flip all logic gates but keep all events in their positive form. To explain why, let us reformulate $X(t)$ as $Y(t)$ where the logical variables are renamed as $y_i = \neg x_i$: 

\vspace{-0.2cm}
\martinListingFontsize\sffamily{ 
    \begin{equation}
    \begin{array}{c}
    Y(t) = (y_1 \lor y_2) \land (y_3 \land y_4 \land (y_5 \lor (y_6 \land y_7))) \nonumber
    \end{array} 
    \vspace{-0.2cm}
    \end{equation}
}\normalsize \normalfont

We know that $\neg Y(t)=f(t)$ by definition. Therefore, we aim at maximising the number of satisfied variables $y_i$ to make $\neg Y(t) = true$. 
But because the variables $y_i$ are the complement of the logical variables $x_i$, we are actually maximising the number of falsified variables $x_i$ and minimising the satisfied ones in $f(t)$. Such a minimal set in $f(t)$ constitutes an MCS in the fault tree. 

\textbf{Step 2 (CNF conversion).}  
SAT solvers normally consider input formulas in conjunctive normal form (CNF). 
To avoid exponential computation times, we use the Tseitin transformation to produce, in polynomial time, a new formula in CNF that is not strictly equivalent to the original formula (because there are new variables) but is equisatisfiable~\cite{Barrere-Arxiv2019}.  
This means that given an assignment of truth values, the new formula is satisfied if and only if the original formula is also satisfied.  

\textbf{Step 3 (Probabilities transformation into log-space).}  
In order to maximise the product of weighted decision variables in MaxSAT, we transform the weights $p(x_i)$ into $w_i = -\log(p(x_i))$ to produce positive values. 
This means that the lower a probability $p(x_i)$, the higher its negative log value $w_i$. 
Conversely, the higher the probability, the lower the $-log$ value, as shown in Table \ref{tab:example-probs} for our example fault tree.  

\bgroup
{    \renewcommand{\arraystretch}{0.8}  
    \setlength{\tabcolsep}{0.4em} 
    \begin{table}[!h]
        \vspace{-0.4cm}
        \centering
        \martinTableFontsize
        \caption{Fault tree probabilities and $-log$ values $w_i$}
        \vspace{-0.3cm}
        \begin{tabular}{c|c|c|c|c|c|c|c}
            \toprule
            \textbf{Probs.} & $x_1$ & $x_2$ & $x_3$ & $x_4$ & $x_5$ & $x_6$ & $x_7$\\
            \midrule
            $p(x_i)$ & 0.2 & 0.1 &  0.001 & 0.002 & 0.05 & 0.1 & 0.05 \\
            $w_i$ & 1.60944 & 2.30259 &  6.90776 & 6.21461 & 2.99573 & 2.30259 & 2.99573 \\		            
            \bottomrule
        \end{tabular}
        \label{tab:example-probs}
    \end{table}
}
\egroup

\vspace{-0.2cm}
\textbf{Step 4 (Weighted Partial MaxSAT instance).}  
We define a soft clause for each decision variable in $\neg Y(t)$. 
These soft clauses indicate the solver that each variable $y_i$ can be falsified with a certain penalty $w_i$, which corresponds to the transformed probability of event $x_i$ as shown in Table~\ref{tab:example-probs}. 
The MaxSAT solver tries to minimise the total weight of falsified variables, and therefore, a solution to this problem yields a minimum vertex cut of the fault tree in logarithmic space. 
Since the lowest logarithmic values correspond to the highest probabilities, the solution indicates the MCS with maximum joint probability, i.e. the \MSCS.  

\textbf{Step 5 (Parallel MaxSAT resolution).}  
We have experimentally observed that, quite often, SAT solvers are very good at some instances and not that good at others. 
This is due to the different optimisation techniques used within solvers \cite{Davies2011}. 
To address this issue, our tool executes multiple pre-configured solvers in parallel and picks up the solution of the solver that finishes first. This method provides a more stable behaviour in terms of performance and scalability. 

\textbf{Step 6 (Reverse log-space transformation).}   
The joint probability of the \MSCS~is computed by performing the reverse log-space transformation $P_F(t) =  \exp(-1 \times \sum_{i} w_i )$, where $i$ indexes the events found in the MaxSAT solution. 

\vspace{-0.05cm}
\section{Preliminary results and conclusion}
\begin{figure}[!t]
    \centering    
    \includegraphics[width=0.41\textwidth]{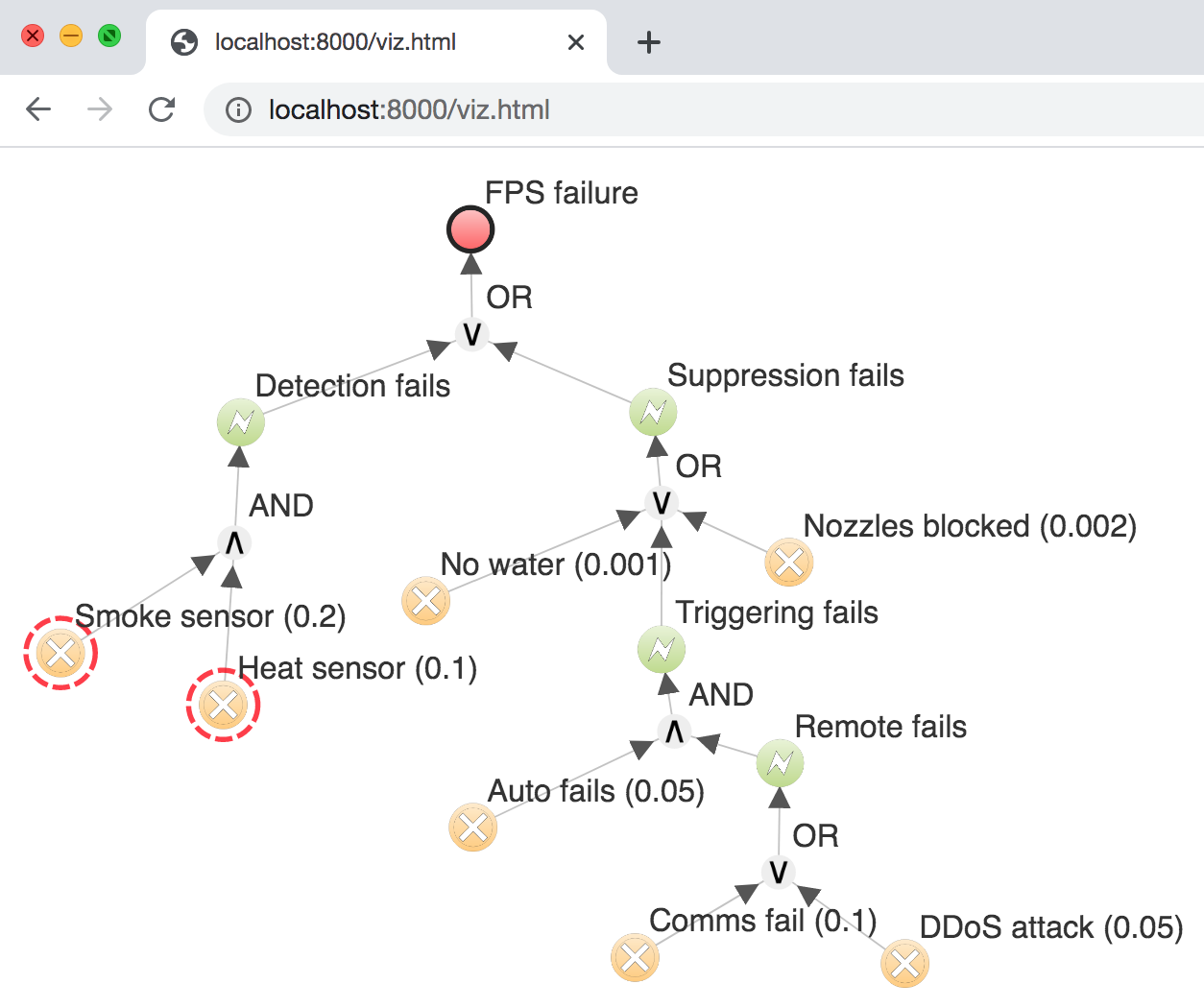}      
    \vspace{-0.35cm}
    \caption{Example scenario and \MSCS~with our tool \toolname} 
    \label{fig:tool-example}
    \vspace{-0.5cm}
\end{figure}

\vspace{-0.05cm}
We have developed an open source tool called \toolname~that implements the proposed methodology and is publicly available at \cite{Barrere-MPMCS4FTA-Github}. 
The tool runs in the command line and outputs the solution in a JSON file that is used to graphically display the fault tree and the \MSCS~in a web browser. 
Fig.~\ref{fig:tool-example} shows the output of \toolname~for our example fault tree. 
The results of our analytical evaluation indicate that the method is able to scale to fault trees with thousands of nodes in seconds. 

FTA is an essential technique to evaluate dependability in a wide range of systems.  
The proposed \MSCS~is intended to extend the body of measures used in FTA and support fundamental activities such as decision making, risk assessment, and fault prioritisation. 
As future work, we plan to evaluate different representation techniques (e.g. BDDs~\cite{Ruijters2015}) to address the \MSCS~problem and conduct a thorough comparison on performance and scalability. We also aim at extending our approach to include additional operators such as voting gates. 

\vspace{-0.1cm}
\bibliographystyle{IEEEtran}
\bibliography{IEEEabrv,doc.bib}

\begin{thebibliography}{1}
\providecommand{\url}[1]{#1}
\csname url@samestyle\endcsname
\providecommand{\newblock}{\relax}
\providecommand{\bibinfo}[2]{#2}
\providecommand{\BIBentrySTDinterwordspacing}{\spaceskip=0pt\relax}
\providecommand{\BIBentryALTinterwordstretchfactor}{4}
\providecommand{\BIBentryALTinterwordspacing}{\spaceskip=\fontdimen2\font plus
\BIBentryALTinterwordstretchfactor\fontdimen3\font minus
  \fontdimen4\font\relax}
\providecommand{\BIBforeignlanguage}[2]{{%
\expandafter\ifx\csname l@#1\endcsname\relax
\typeout{** WARNING: IEEEtran.bst: No hyphenation pattern has been}%
\typeout{** loaded for the language `#1'. Using the pattern for}%
\typeout{** the default language instead.}%
\else
\language=\csname l@#1\endcsname
\fi
#2}}
\providecommand{\BIBdecl}{\relax}
\BIBdecl

\bibitem{FtaHandbook2002}
W.~Vesely, M.~Stamatelatos, J.~Dugan, J.~Fragola, J.~Minarick~III, and
  J.~Railsback, ``{Fault Tree Handbook with Aerospace Applications},''
  \emph{Office of Safety and Mission Assurance, NASA Headquarters, US}, 2002.

\bibitem{Ruijters2015}
E.~Ruijters and M.~Stoelinga, ``Fault tree analysis: A survey of the
  state-of-the-art in modeling, analysis and tools,'' \emph{Computer Science
  Review}, vol. 15-16, pp. 29 -- 62, 2015.

\bibitem{Kabir2017}
S.~Kabir, ``{An overview of Fault Tree Analysis and its application in model
  based dependability analysis},'' \emph{Expert Systems with Applications},
  vol.~77, pp. 114 -- 135, 2017.

\bibitem{Barrere-Arxiv2019}
M.~Barr\`ere, C.~Hankin, N.~Nicolaou, D.~Eliades, and T.~Parisini,
  ``{Identifying Security-Critical Cyber-Physical Components in Industrial
  Control Systems},'' \url{https://arxiv.org/abs/1905.04796}, May 2019.

\bibitem{Davies2011}
J.~Davies and F.~Bacchus, ``{Solving MAXSAT by Solving a Sequence of Simpler
  SAT Instances},'' in \emph{Principles and Practice of Constraint Programming
  -- CP 2011}, J.~Lee, Ed.\hskip 1em plus 0.5em minus 0.4em\relax Springer,
  2011, pp. 225--239.

\bibitem{Barrere-MPMCS4FTA-Github}
M.~Barr\`ere, ``{MPMCS4FTA - Maximum Probability Minimal Cut Sets for Fault
  Tree Analysis},'' \url{https://github.com/mbarrere/mpmcs4fta}, Mar. 2020.

\end{thebibliography}

\end{document}